\def\year{2020}\relax

\documentclass[letterpaper]{article} 
\usepackage{aaai20}  
\usepackage{times}  
\usepackage{helvet} 
\usepackage{courier}  
\usepackage[hyphens]{url}  
\usepackage{graphicx} 
\usepackage{graphicx}  
\usepackage{algorithm}
\usepackage{algpseudocode}
\usepackage{amsmath}
\usepackage{amssymb}
\usepackage{mathtools}
\usepackage{color}

\title{Learning Neural Search Policies for Classical Planning}
\author{
Pawe\l{} Gomoluch,\textsuperscript{\rm 1} Dalal Alrajeh,\textsuperscript{\rm 1} Alessandra Russo,\textsuperscript{\rm 1} Antonio Bucchiarone\textsuperscript{\rm 2}\\
\textsuperscript{\rm 1}Department of Computing, Imperial College London, \textsuperscript{\rm 2}Fondazione Bruno Kessler, Trento, Italy\\
\{pawel.gomoluch14,dalal.alrajeh,a.russo\}@imperial.ac.uk, bucchiarone@fbk.eu
}

\begin{document}

\maketitle

\begin{abstract}
Heuristic forward search is currently the dominant paradigm in classical planning.
Forward search algorithms typically rely on a single, relatively simple variation of best-first search and remain fixed throughout the process of solving a planning problem.
Existing work combining multiple search techniques usually aims at supporting best-first search with an additional exploratory mechanism, triggered using a handcrafted criterion.
A notable exception is very recent work which combines various search techniques using a trainable policy. It is, however, confined to a discrete action space comprising several fixed subroutines.

In this paper, we introduce a parametrized search algorithm template which combines various search techniques within a single routine. The template's parameter space defines an infinite space of search algorithms, including, among others, BFS, local and random search. We further introduce a neural architecture for designating the values of the search parameters given the state of the search. This enables expressing neural search policies that change the values of the parameters as the search progresses. The policies can be learned automatically, with the objective of maximizing the planner's performance on a given distribution of planning problems. We consider a training setting based on a stochastic optimization algorithm known as the \emph{cross-entropy method} (CEM).
Experimental evaluation of our approach shows that it is capable of finding effective distribution-specific search policies, outperforming the relevant baselines.
\end{abstract}

\section{Introduction}

Modern classical planners usually rely on heuristic forward search. Much research effort in recent years has been devoted to the development of advanced domain-independent heuristic functions (e.g. \cite{Hoffmann2001,Helmert2006,Richter2010,Domshlak2015}). In contrast, the search algorithms at the core of many successful planners have largely remained simple variations of best-first search, such as greedy best-first search \cite{Helmert2006} or weighted A* \cite{Richter2010}. In recent years, some work has sought to combine best-first search with additional exploratory mechanisms, such as randomized order of node expansion \cite{Valenzano2014,Asai2017}, random walks and local search \cite{Xie2014} or novelty search \cite{Lipovetzky2017}. The motivation behind these approaches is to enable the search to escape local minima and plateaus of the heuristic function.

Introducing even a single auxiliary exploration mechanism necessarily comes with a number of nontrivial design choices, such as when to switch between the main and auxiliary search approach. In addition, many of the exploration mechanisms come with a number of parameters of their own, such as the length and number of random walks to perform. The values of the parameters are typically selected by human experts. Furthermore, they stay fixed throughout the process of solving the problem.

The work introduced in this paper aims at automating the design of multi-technique search algorithms. To achieve it, we first construct a parametrized search algorithm which combines multiple search techniques in a flexible manner. Depending on the values of the parameters, the search algorithm can take form of BFS, iterated local search or random search, among others. Rather than choosing fixed values of the parameters, we introduce a trainable model mapping the current state of the search to an assignment over the parameters. We then train the model using the cross-entropy method (CEM) to obtain search policies tailored to specific problem distributions. To the best of our knowledge no existing work has focused on a similar problem, except a very recent one \cite{Gomoluch2019}. That approach, however, is limited to selection from a discrete set of several fixed search routines. We discuss it in more detail in the following section.

We implement our approach within the Fast Downward planning system \cite{Helmert2006} and evaluate it using five domains from the \emph{International Planning Competition} (IPC). The learned search policies outperform baselines built around the component techniques, as well as a fixed hand-crafted combination all the techniques available to the parametrized planner.

\section{Related Work}

By using learning to improve the performance of a planner, our work joins the ranks of numerous learning approaches in classical planning. Over the years, learning has been used to acquire macro operators \cite{Fikes1972,Coles2007,Gerevini2009}, heuristic functions \cite{Yoon2008,Virseda2013,Garrett2016} and search control rules \cite{Leckie1998,Yoon2008}.

Little work has sought to learn directly in the space of possible search algorithms until very recently, when policy gradient was used to learn search strategies \cite{Gomoluch2019}. That work is closely related to ours in combining various search techniques in a single search run and in learning how to best do so, given a particular distribution of planning problems. The most important difference is the space of possible search policies. In that work, a policy is trained to choose a search subroutine out of a set of five. While shown to be effective, this approach imposes somewhat arbitrary restrictions on the action and policy spaces. For example, the planner can choose to perform $\epsilon$-greedy search, but not the value of epsilon. It can choose to perform random walks but not set their length. The span of local search is also effectively determined by one of the hyperparameters. Further, local and $\epsilon$-greedy search are made mutually exclusive, although in some domains it may be beneficial to combine the two, i.e. randomize the order of node expansion in local search. In our approach, various search techniques are merged into a single parametrized search routine that subsumes each of the subroutines on its own for some specific assignment of the parameters. In reinforcement learning terms, the previous work deals with a discrete action space while our is close in spirit to continuous action spaces.\footnote{The action space is not strictly continuous as some of the search parameters take integer values. However, the presence of real-valued parameters makes the space infinite.}

Another important difference is the representation of the search policies. The existing work \cite{Gomoluch2019} adopts a tabular approach with four states generated by two binary features. In our work, we use numeric features and a function approximator mapping them into the values of search parameters.

A different approach to planning with multiple search techniques is to divide the available time and perform a number of separate search runs. The original Fast Forward planner \cite{Hoffmann2001} first attempts to solve the problem using \emph{enforced hill climbing}. If this fails, it falls back to best-first search. The LAMA planner \cite{Richter2010} follows greedy search with a number of weighted A* runs with decreasing weights in order to find the first solution as quickly as possible and then keep improving on it while the time permits. The idea is taken even further in the planner portfolios which run a number of potentially unrelated solvers independently (e.g. \cite{Fawcett2011,Helmert2011,Cenamor2016}). Our approach differs from the portfolios by dealing with a single search run and learning how to best proceed with the current search frontier. In doing so, it is complementary to portfolio approaches. For example, future work could consider building portfolios composed of parametrized planners optimized for various problem distributions.

\section{Background}

\subsubsection{Classical Planning} Planning is the problem of finding sequences of actions leading from an given initial state to state in which a given goal is satisfied. Classical planning in particular relies on a perfect and deterministic world model known to the planner. Formally, the classical planning task is given by a tuple $\langle V,O,S_0,g\rangle$, where $V$ is a set of finite-domain variables, $O$ is the set of operators, $s_0$ is the initial state and $g$ is the goal. The initial state $s_0$ is an assignment over the variables of $V$. The goal $g$ is a partial assignment over $V$. The operators $o \in O$ are themselves defined by specifying their preconditions $\operatorname{pre}(o)$ and effects $\operatorname{eff}(o)$, both of which are partial assignments over $V$. Operator $o$ can be applied in state $s$ if and only if its preconditions are satisfied in the state, $\operatorname{pre}(o) \subseteq s$. The state resulting from applying an operator $o$ to state $s$, $o(s)$ is determined by setting the values of variables covered by $\operatorname{eff}(o)$ to the corresponding values, and keeping all the other variables unchanged. The task is to find a sequence of operators $o_0, o_1, o_n$, such that applying them in order leads to satisfaction of the goal $g$: $g \subseteq o_n(o_{n-1}(...o_0(s_0)...))$.

The most common approach to classical planning is forward search. Forward search starts by inserting a node containing the initial state $s_0$ in the \emph{open list}. It proceeds by iteratively removing nodes from the list and expanding them by applying all of the operators applicable in the given state and adding the resulting nodes to the open list. This is repeated until a state satisfying the goal $g$ is reached or the open list becomes empty, which indicates the the problem is unsolvable. Best-first search (BFS) always expands the node $n$ with the lowest value of some evaluation function $f(n)$. If $f(n)$ depends solely on the heuristic estimate of the distance to the goal $f(n)=h(n)$ the search becomes \emph{greedy} best-first search (GBFS).

\subsubsection{Cross-Entropy Method} The cross-entropy method is a gradient-free stochastic optimization technique originating from rare event simulation \cite{Rubinstein1999,DeBoer2005}. It belongs to a wider family of population-based optimization techniques known as Evolution Strategies (ES), which also includes approaches such as Covariance Matrix Adaptation \cite{Hansen2O01} and Natural Evolution Strategies \cite{Wierstra2014}. In recent years, various forms of evolution strategies have been successfully applied as policy search methods in a number of challenging reinforcement learning domains \cite{Mannor2003,Salimans2017,Chrabaszcz2018,Conti2018}. The common idea underlying ES is to maintain a population of candidate solutions and iteratively update it towards solutions of higher quality.

Given a (possibly stochastic) function $s(\mathbf{x}), \mathbf{x} \in \mathcal{X}$, the cross-entropy method introduces an auxiliary distribution $f_v(\mathbf{x})$ over the possible solutions. The distribution is itself parametrized by a vector $\mathbf{v}$. For example if the distribution is a multivariate Gaussian, $v$ can contain its mean and flattened covariance matrix. At each iteration $t$, $n$ candidate solutions $\mathbf{x}^{(1)}_t, \mathbf{x}^{(2)}_t \ldots \mathbf{x}^{(n)}_t$ are sampled from the distribution and evaluated. The parameters of the distribution are then updated to maximize the likelihood of the $m$ best candidate solutions ($m<n$). To select the best solutions, a performance threshold $\gamma_t$ is introduced, equal to the $\frac{m}{n}\cdot100$-th percentile of the candidate scores $s(\mathbf{x}^{(1)}_t), s(\mathbf{x}^{(2)}_t) \ldots s(\mathbf{x}^{(n)}_t)$. The value of $\mathbf{v}$ for the next iteration is designated by solving:
\begin{equation}
\label{eq:cem}
\mathbf{\tilde{v}}_{t} = \underset{\mathbf{v} \in \mathcal{V}}{\operatorname{argmax}} \frac{1}{N} \displaystyle{\sum_{i=1}^{N}}\mathbf{I}_{s(\mathbf{x}^{(i)}_t)>\gamma_{t}} \log f(\mathbf{x}^{(i)}, \textbf{v}_t)
\end{equation}
where $ \mathbf{I}_{s(\mathbf{x}^{(i)}_t)>\gamma_{t}} $ is the indicator variable of the event $ s(\mathbf{x}^{(i)}_t)>\gamma_{t} $, i.e. it is 1 if the score of sample $i$ is higher than the threshold $\gamma_t$ and 0 otherwise.

In the particular case when the auxiliary distribution is a multivariate Gaussian, solving the Equation \ref{eq:cem} amounts to setting $\mathbf{\tilde{v}}_{t}$ to the mean and covariance of the $m$ samples above the threshold $\gamma_t$.

Instead of using the solution of Equation \ref{eq:cem} directly for the next iteration's distribution ($\mathbf{v}_{t+1} = \mathbf{\tilde{v}_t}$), it is also possible to perform a \emph{smoothed} update:
\begin{equation}
    \mathbf{v_{t+1}} = \alpha{ \mathbf{\tilde{v}_t}} + (1-\alpha)\mathbf{v_t}
\end{equation}

The process can be repeated for a fixed number of iterations or until a specific stopping criterion is met. After the final iteration $n$ the result can be obtained by extracting the mean of the final distribution from $\mathbf{v_{t+1}}$.

\section{Parametrized Search} \label{sec:parametrized-search}


The key idea underlying this work is that multiple forward search techniques can be combined in a single forward search algorithm. For instance, the algorithm can interleave between global and local best first search by performing a number of standard node expansions and then a number of local expansions. Both local and global expansions can optionally be followed by a number of random walks started from the expanded node. Additionally, both global and local search can randomize the order of node expansion, by choosing a random node from the open list instead of the one with lowest $h$, with the probability of $\epsilon$.

Pseudocode implementing this idea is presented in Algorithm \ref{alg:pp}. Typically for a forward approach, the search starts by initializing the open list to contain the initial state $s_0$ (line \ref{a1:init}). The main loop of the algorithm (lines \ref{a1:main-while}-\ref{a1:end-main-while}) performs a number of search steps using the global open list (lines \ref{a1:global-for}-\ref{a1:end-global-for}). Then it initializes a new local open list with a single node popped from the global list (line \ref{a1:local-init}). The local list is used to perform a number of local search steps (lines \ref{a1:local-for}-\ref{a1:end-local-for}). After the local search, all the nodes from the local list are merged into the global one (line \ref{a1:merge}). The number of global and local search steps in a single iteration of the main loop is controlled by two parameters: $C$ is the total number of steps while $c$ is the proportion of local steps. One iteration of the main loop consists of $(1-c) \cdot C$ global and $c \cdot C$ local steps.

A search step (lines \ref{a1:step}-\ref{a1:end-step}) involves a single node expansion (lines \ref{a1:expansion}-\ref{a1:end-expansion}), optionally followed by $N$ random walks of length $L$, starting at the expanded node (lines \ref{a1:random-walk}-\ref{a1:end-random-walk}). One of the details omitted from Algorithm \ref{alg:pp} is keeping track of the lowest known heuristic value $h_{\mathrm{min}}$. The value is initialized to the heuristic evaluation of the initial state and then updated whenever a state with lower evaluation is encountered. Similarly, the algorithm needs to keep track of the number of node expansions performed since the last update of $h_{\mathrm{min}}$. This value is used to decide whether to follow the current node expansion with random walks (line \ref{a1:random-walk}). The walks are triggered on condition that no decrease of $h_{\mathrm{min}}$ has been observed during the last $S$ expansions. The $S$ parameter plays a role analogous to the \emph{STALL\_SIZE} parameter used by \cite{Xie2014} to decide when to switch from GBFS to the auxiliary exploration technique.

If the step is using the local list and it turns out to be empty, another node is moved from the global list to the local one (line \ref{a1:reseed}). If the global list becomes empty without reaching the goal, the search fails (lines \ref{a1:fail1} and \ref{a1:fail2}).

Overall, the search parameters include:
\begin{itemize}
    \item $\epsilon$ -- the probability of selecting a random node from the open list;
    \item $S$ -- the number of expansions without progress necessary to trigger a random walk;
    \item $R$ -- the number of random walks following a single node expansion;
    \item $L$ -- the length of a random walk;
    \item $C$ -- the number of node expansions in the global-local cycle;
    \item $c$ -- the proportion of local search in the global-local cycle.
\end{itemize}

Assigning the values of the parameters positions the resulting algorithm in the space between various search approaches. For example, with $\epsilon=0$, $R=0$ and $c=0$ the algorithm becomes greedy BFS, independently of the remaining parameters. With $\epsilon=0$, $R=0$, $c=1$ and $C=100$, the algorithm performs iterated local search with a span of 100. Intermediate values of $c$ result in interleaving of global and local search, with random walks optionally added in both phases, as determined by remaining parameters.

The values of the search parameters can be changed during execution of the algorithm. It is convenient to update them at the beginning of every iteration of the main loop (line \ref{a1:set-parameters}) and keep them fixed throughout the iteration. In the following sections we introduce the task of learning search policies, whose purpose is to set the values of the parameters, given the current state of the search. 

\begin{algorithm}[ht!]
\caption{Parametrized planner}
\label{alg:pp}
\begin{algorithmic}[1]
\Function{Planner}{$s_0, g, O$}
\State $global\_open \gets [s_0]$ \label{a1:init}
\While{true} \label{a1:main-while}
\State $\epsilon, S, R, L, C, c \gets \operatorname{set\_search\_parameters()}$ \label{a1:set-parameters}
\For{$i = 1 \ldots (1-c) \cdot C$} \label{a1:global-for}
    \State \Call{Step}{$global\_open$}
\EndFor \label{a1:end-global-for}
\State $local\_open \gets [\operatorname{pop}(global\_open)]$ \label{a1:seed-local}
\For{$i = 1 \ldots c \cdot C$} \label{a1:local-for}
    \State \Call{Step}{$local\_open$} \label{a1:local-init}
\EndFor \label{a1:end-local-for}
\State merge $local\_open$ into $global\_open$ \label{a1:merge}
\EndWhile \label{a1:end-main-while}
\EndFunction
\State
\Function{Step}{$open, O, g$} \label{a1:step}
\If{$open$ is empty}
    \If{$open \operatorname{is} global\_open$}
        \State return failure \Comment{return from \textproc{Planner}} \label{a1:fail1}
    \Else
        \If{$global\_open$ is empty}
            \State return failure \Comment{return from \textproc{Planner}} \label{a1:fail2}
        \Else
            \State $local\_open \gets [\operatorname{pop}(global\_open)]$ \label{a1:reseed}
        \EndIf
    \EndIf
\EndIf
\State $s \gets \operatorname{pop}(open, \epsilon)$ \label{a1:expansion}
\If{$s \in g$}
    \State $plan \gets \operatorname{extract\_plan}(s)$
    \State return $plan$ \Comment{return from \textproc{Planner}}
\EndIf
\State $successor\_states \gets \operatorname{expand}(s, O)$
\State $\operatorname{add}(open, successor\_states)$ \label{a1:end-expansion}
\If{$expansions\_without\_progress > S$} \label{a1:stall} \label{a1:random-walk}
\For{$i = 1 \ldots R$} \label{a1:random-for}
    \State $walk\_states \gets \operatorname{random\_walk}(s, L)$
    \State $\operatorname{add}(open, walk\_states)$
\EndFor \label{a1:end-random-for}
\EndIf \label{a1:end-random-walk}
\State return in\_progress \Comment{return to the \textproc{Planner} loop}
\EndFunction \label{a1:end-step}
\end{algorithmic}
\end{algorithm}

\section{Search State Representation} \label{sec:representation}

To facilitate learning of state-dependent search policies, we introduce a high-level representation of the state of the planner. In particular, we consider the following features of the planner's state:
\begin{itemize}
    \item the heuristic value of the initial state $h(s_0)$;
    \item the lowest heuristic value encountered within the search $h_{\mathrm{min}}$;
    \item the time elapsed since the search started;
    \item the number of node expansions performed since the last change in the value of $h_{\mathrm{min}}$;
    \item the number of states the search has generated;
    \item the number of unique states the search has generated;
    \item the total number of nodes the search has expanded.
\end{itemize}
Intuitively, the features capture important information about the state of the search. For example, a large number of node expansions performed since the last decrease in $h_{\mathrm{min}}$ indicates that the planner is facing a local minimum or a plateau of the heuristic function. In such a situation, increasing the amount of exploratory behavior is likely to be beneficial. Comparing the current value of the heuristic with $h(s_0)$ enables estimating the relative progress towards the goal. If the values are close, the search has likely made little progress towards the goal. On the other hand, a value of $h_{\mathrm{min}}$ close to 0 can suggest that the search frontier is close to the goal. Naturally, the reliability of such estimates will depend on how well the heuristic function correlates with actual distance to the goal in the current domain. This information can be particularly useful in conjunction with the elapsed time $t$. If sufficient progress is being made at early stage of the search, there may be no incentive to deploy techniques making rapid progress towards the goal at the cost of sacrificing the plan quality (for example random walks in logistics-style domains). On the other hand, if time is running out while significant distance remains to be covered, more aggressive approach may be desirable in order to avoid failure due timeout.

In the following section, these features are used as the state representation for learning search policies.

\section{Learning Search Policies}

We consider a search policy to be a vector valued function mapping the state of the search to the values of the search parameters. Formally, $Y=\pi(\Phi)$, where $Y$ is a vector of the search parameters $\langle \epsilon, S, R, L, C, c\rangle$, introduced in Section \ref{sec:parametrized-search}, and $\Phi$ is the vector of state features listed in Section \ref{sec:representation}.

\subsection{Policy model}

We represent the search policies using a parametric function approximator $\pi_{\theta}(\Phi)$, with trainable parameters $\theta$.
Specifically, we use a feed-forward neural network. The inputs of the network are the planner's state features and its outputs are the search parameters. The network contains one hidden layer of 7 units with sigmoid activation function $f(z)=\frac{1}{1+e^{-z}}$. The treatment of the network's output depends on the search parameter it represents. For real-valued search parameters ($\epsilon$ and $c$) taking values from range $(0,1)$, the network outputs are passed through a sigmoid function. For search parameters taking nonnegative integer values ($S$, $R$, $L$ and $C$) the outputs are restricted to nonnegative values $y_r=\operatorname{max}(y,0)$ and truncated to the integer part.

Because the planner's state features differ in order of magnitude, we scale them before passing them through the network. We choose the scaling factors on a per-domain basis, by running our parametrized planner with fixed parameter values on a separate batch of problems from the domain and recording the maximum values $\phi_{\mathrm{max}}$ of all the features observed in the process. The maximum values are then used to scale each of the features $\phi_{\mathrm{scaled}}=\frac{\phi}{\phi_{\mathrm{max}}}$, so that value 1 of the scaled feature represents the highest value recorded on the batch of problems. The scales remain unchanged throughout training and testing of the system.

Similarly, we scale the outputs of the network. With the exception of $\epsilon$ and $c$, whose values are put in the range of $(0,1)$ by the sigmoid, we scale the outputs so that the value of 1 results in moderate use of the corresponding search technique. Concretely, we multiply the number of random walks $R$ by 5, the length of the walks $L$ by 10, the stall condition $S$ by 10 and the length of the global-local cycle $C$ by 100.

\subsection{Policy evaluation}

To learn search policies in an evolutionary process, it is necessary to establish a way of evaluating candidate policies. In this work, we adopt an approach based on the scoring function used in the satisficing track of the \emph{International Planning Competition} (IPC) since the 2008 edition\footnote{\url{http://icaps-conference.org/ipc2008/deterministic/}}. For a failed problem, a planner receives a score of 0. For a solved one the score is defined as:
$$ g = \frac{c_{\textrm{min}}}{c}$$
where $c$ is the cost of the found plan, and $c_\textrm{min}$ is the lowest known cost of a plan solving the problem. In practice, $c_{min}$ is often the lowest cost of a plan returned by any of the competing planners. The score obtained on a set of problems is the sum of scores for each of the problems.
$$ G = \displaystyle{\sum_p}g_p$$

 In our training setting, detailed in the next section, the candidate policies are put in competition against each other: $c_{\mathrm{min}}$ is the lowest solution cost found by any of the candidate policies. This is convenient since the problems are generated randomly for each iteration, and so no reference costs are known a priori. We remark that when training on a fixed set of problems, the lowest plan cost could be retained between iterations or designated using a set of reference planners.

\subsection{Policy learning}

To learn search policies best suited to specific problem distributions, we employ the cross-entropy method \cite{Rubinstein1999,Mannor2003}. We introduce a multivariate Gaussian distribution over the policy parameters $\theta$, itself parametrized by the mean vector $\mu$ and covariance matrix $\Sigma$.
$$\theta \sim \mathcal{N}(\mu, \Sigma)$$
We initialize $\mu$ with a zero vector and $\Sigma$ with an identity matrix.

At each iteration $i$, $n$ samples $\theta_1$...$\theta_n$ are drawn from $\mathcal{N}(\mu_i, \Sigma)$. The resulting policies are evaluated on set of training problems. The mean and covariance of the parameter distribution are then updated towards the mean and covariance of the $m$ candidate solutions with the best performance:
\begin{equation}
\nu_t = \displaystyle\sum_{i=1}^{m}\theta_i^{t}
\end{equation}
\begin{equation}
\mu_{t+1} = (1-\alpha)\mu_t + \alpha\nu_t
\label{eq:mu}
\end{equation}
\begin{equation}
\Sigma_{t+1} = (1-\alpha)\Sigma_t + \alpha\frac{1}{m-1}\displaystyle\sum_{i=1}^{m}(\theta_i^{t}-\nu_{t})(\theta_i^{t}-\nu_{t})^\top
\label{eq:sigma}
\end{equation}
where $\alpha$ is a smoothing factor and $\theta_1^{t}$...$\theta_m^{t}$ are the $m$ best scoring candidates of iteration $t$.

The pseudocode of the learning approach is presented in Algorithm \ref{alg:learning}. The inputs of the algorithm include a distribution of planning problems ($\mathcal{P}$), the number of iterations ($u$), the number of planning problems sampled at each iteration ($r$), the number of policies sampled at each iteration ($n$) and the number of elite policies used to update the policy distribution ($m$).

The main loop of the algorithm starts by randomly generating a set of problems following the target distribution (line \ref{a2:problem-generation}). Generally, the problems $p_1$...$p_r$ do not have to be independently and identically distributed. As detailed in Section \ref{sec:experiments}, in our experiments we use problem distributions resembling the IPC problem sets, which contain problems of varying difficulty. In the absence of problem generators, this step could be replaced by sampling from a fixed set of training problems.

Further, $n$ policies are sampled from the current policy distribution (line \ref{a2:sampling}). Each of the policies is used to tackle each of the generated problems (lines \ref{a2:for}-\ref{a2:end-for}). The results are used to compute IPC score for each policy (line \ref{a2:scores}). The policies are then sorted according to their scores in order to select the $n$ best performing ones. The new mean of the policy distribution is obtained by averaging the $n$ samples and performing a weighted sum with the old mean (line \ref{a2:update}). The algorithm returns the final mean of the policy distribution as the resulting policy (line \ref{a2:return}).

\begin{algorithm}
\caption{Policy learning} 
\label{alg:learning}
\begin{algorithmic}[1]
\Function{train}{$\mathcal{P}, u, r, n, m$}
\State initialize $\mu$ and $\Sigma$ 
\For {$i = 1 ... u$}
    \State $p_1 ... p_r \gets \mathcal{P}$ \Comment{sample $r$ problems} \label{a2:problem-generation}
    \State $\theta_1 ... \theta_n \gets \mathcal{N}(\mu, \Sigma)$ \Comment{sample $n$ policies} \label{a2:sampling}
    \For {$j = 1 ... n$} \label{a2:for}
        \For {$k = 1 ... r$}
            \State run policy $\theta_j$ on $p_k$, record plan cost $c_{j,k}$
        \EndFor
    \EndFor \label{a2:end-for}
    \State $G_1...G_n \gets$ compute IPC score for $\theta_1 ... \theta_n$ \label{a2:scores}
    \State sort $\theta_1...\theta_n$ by scores $G_1...G_n$ (highest first)
    \State update $\mu$ and $\Sigma$ according to Equations \ref{eq:mu} and \ref{eq:sigma} \label{a2:update}
\EndFor
\State return $\mu$ \label{a2:return}
\EndFunction
\end{algorithmic}
\end{algorithm}

\section{Experiments} \label{sec:experiments}

We have implemented the parametrized planner as a component of the Fast Downward \cite{Helmert2006} planning system. The source code of the planner and learner is available online\footnote{URL removed for blind review}.

We evaluate our approach on five IPC domains: \emph{Elevators}, \emph{Floortile}, \emph{No-mystery}, \emph{Parking} and \emph{Transport}. These are all the domains of the learning track of IPC 2014, with the exception of the \emph{Spanner} domain. The latter was designed specifically not to work well with delete-relaxation heuristics, such as the Fast Forward heuristic \cite{Hoffmann2001}, which we use throughout our experiments. For problem generation, we use the problem generators published by the organisers of the learning track\footnote{\url{http://www.cs.colostate.edu/~ipc2014/}}. In terms of problem size, we use problem distributions of the satisficing track of IPC 2011, which is the last edition in which the domains occurred together. Note that by a problem distribution we mean the values of the parameters passed to the problem generators, which do not necessarily give full account of the difficulty of the actual IPC 2011 problem sets. This is because the selection of the competition problems may have involved additional criteria not captured by the generator parameters, for example manual rejection of problems that have empirically proven too easy or too hard for some set of reference planners. For both training and evaluation we use a time limit of 3 minutes per problem, which is lower than the 30 minutes traditionally used in IPC. The primary motivation for this is the training setting, which requires multiple planner runs at every iteration. We remark that training with a larger time limit is equally possible, although significantly more expensive, as it requires up to 10 times more computation time.

\subsection{Training}

We train a separate model for each of the domains. At every iteration, we randomly generate 20 problems of difficulty roughly corresponding to the problem sets of the satisficing track of IPC. Every batch of randomly generated problems preserves the generator parameters used for the competition set. For example, if the \emph{Parking} set of the 2011 competition contains 20 problems including two problems with 22, three problems with 24 cars, three problems with 26 cars and so on, then the same holds for every of the training batches.

\begin{figure*}
    \centering
    \begin{tabular}{cc}
         \includegraphics[width=0.45\textwidth]{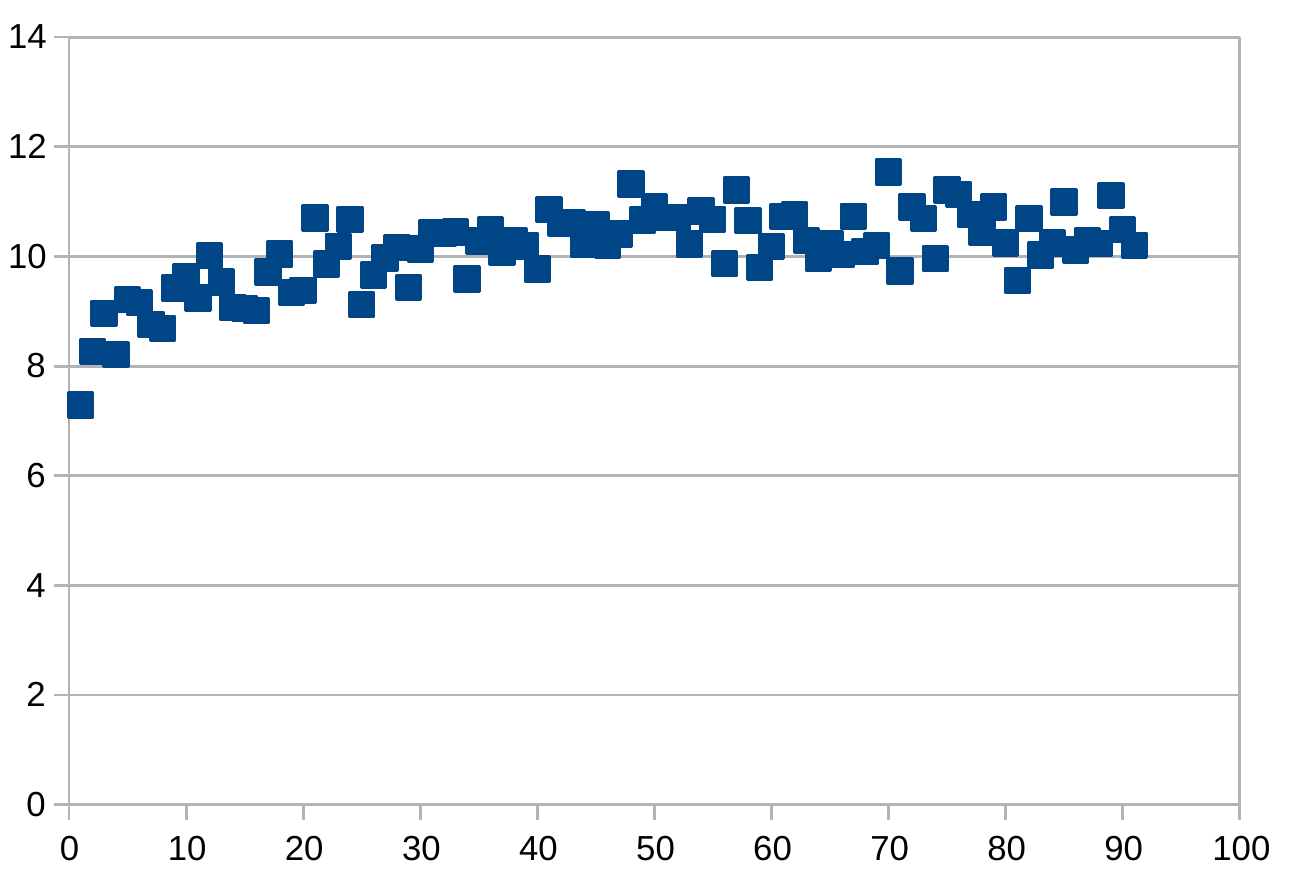} & \includegraphics[width=0.45\textwidth]{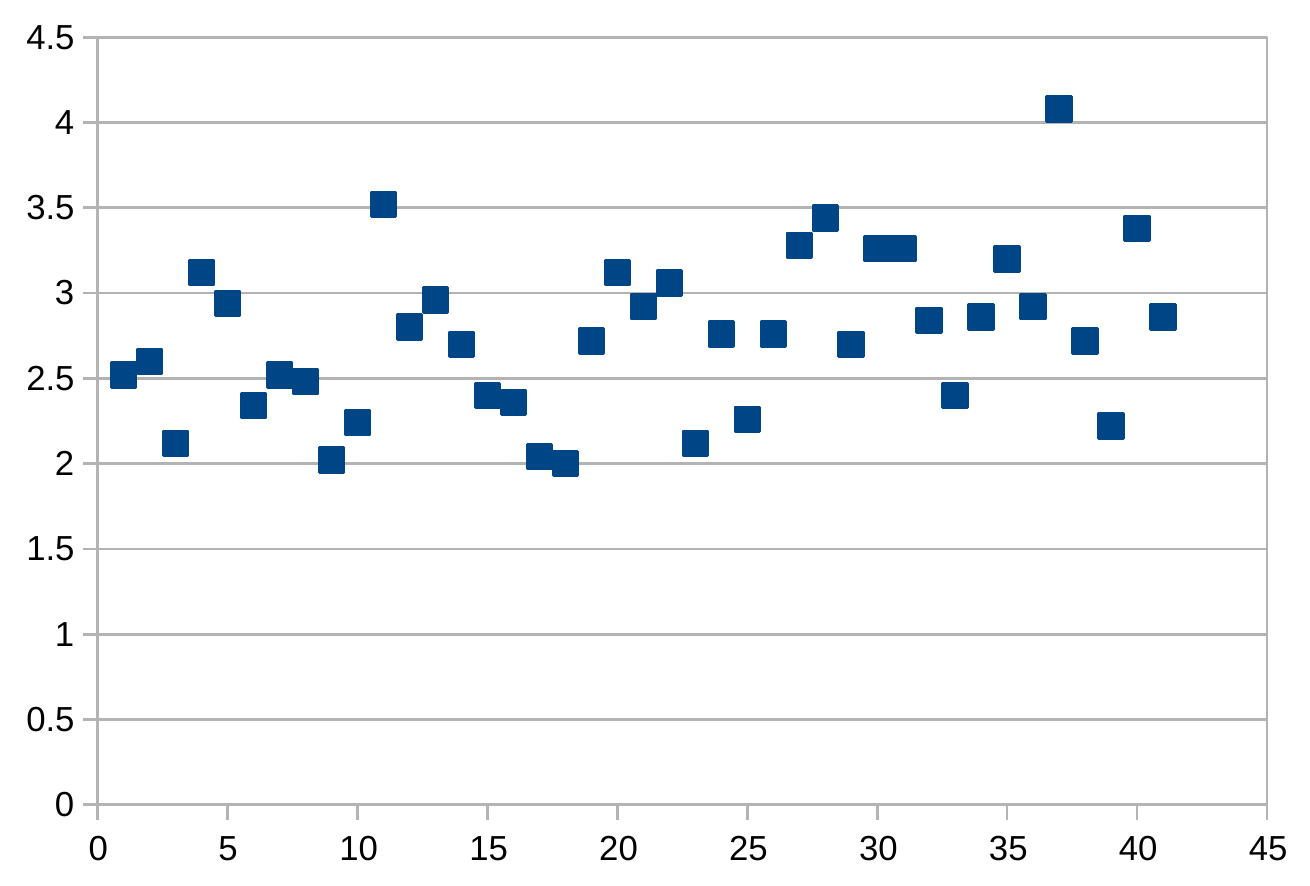} \\
         Elevators\vspace{2ex} & Floortile \\
         \includegraphics[width=0.45\textwidth]{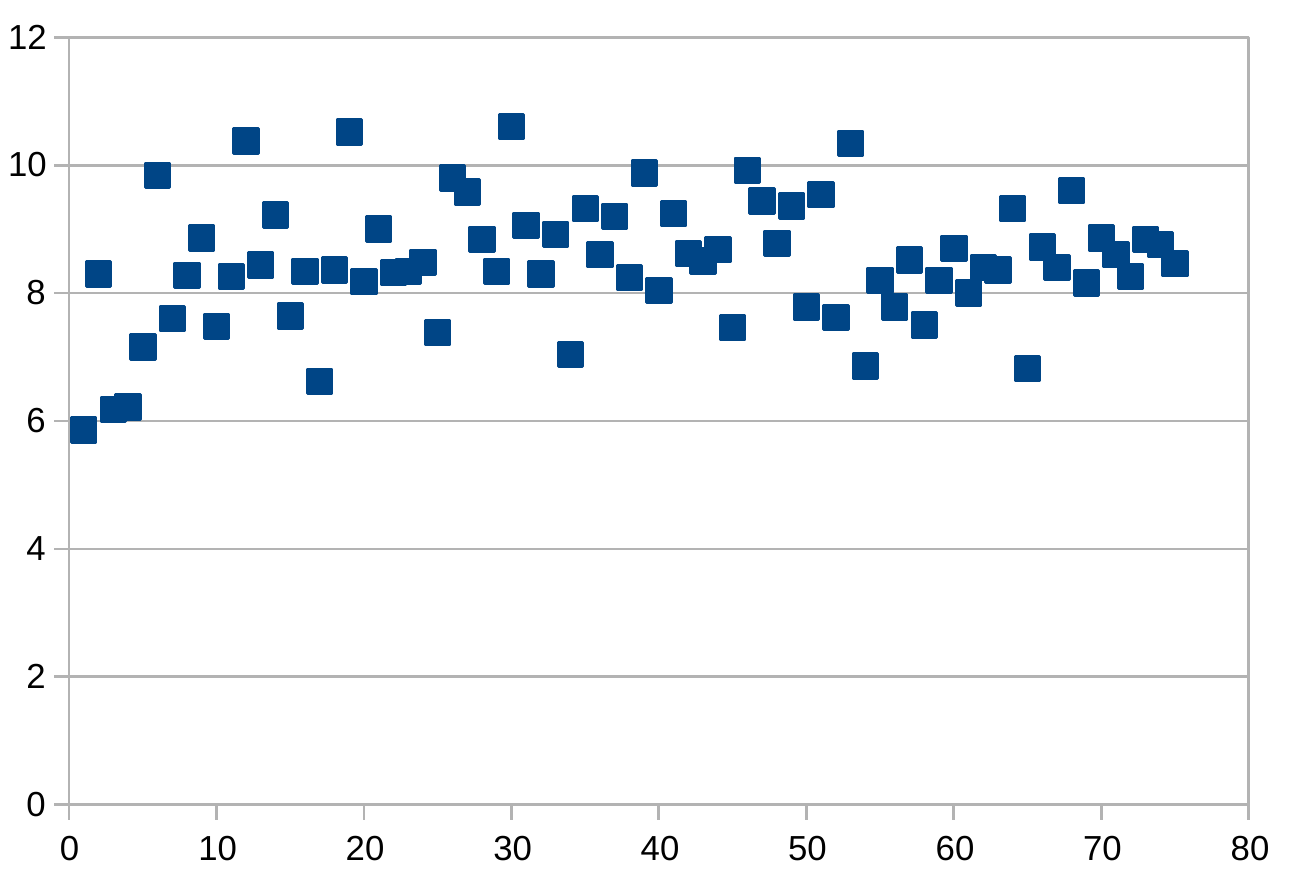} &
         \includegraphics[width=0.45\textwidth]{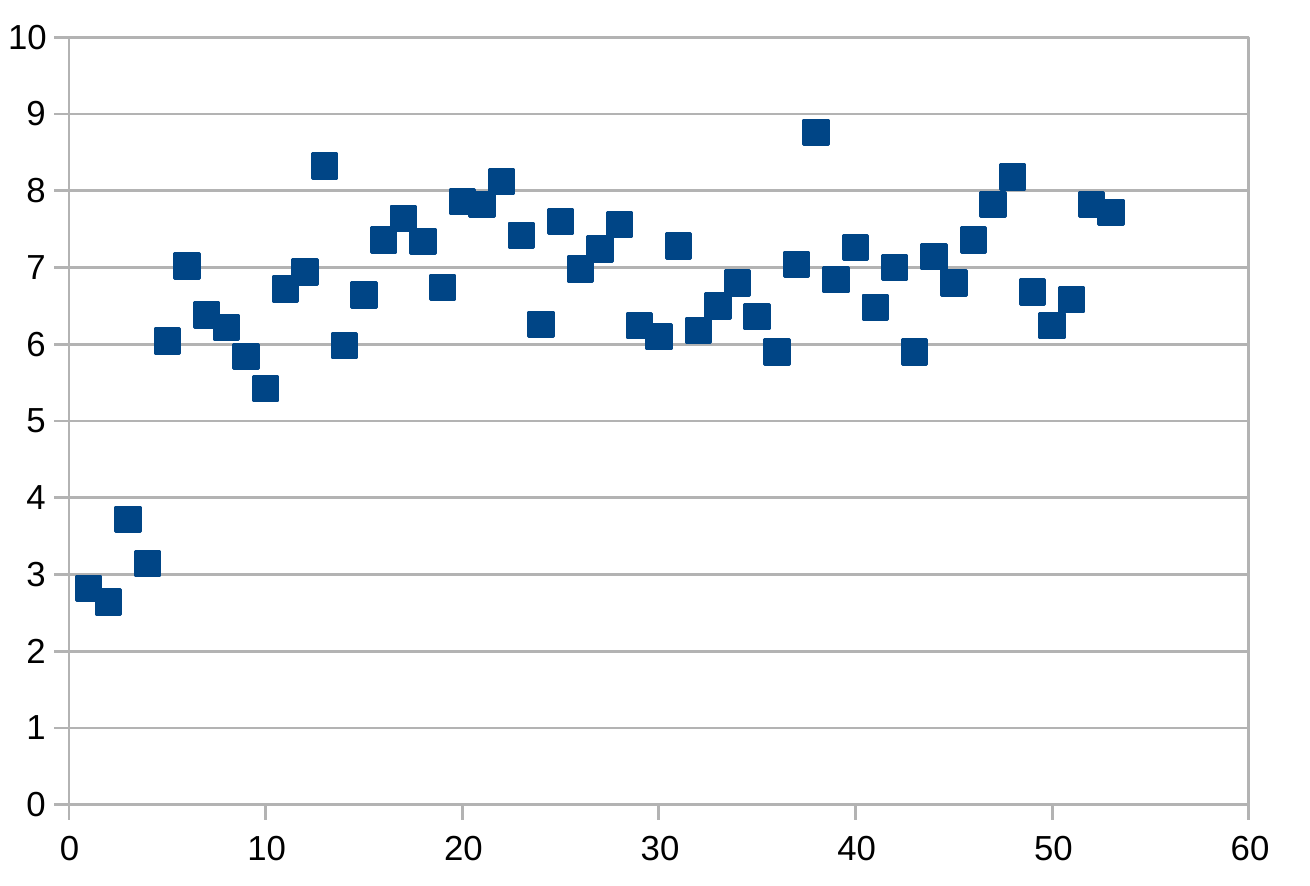} \\
         No-mystery\vspace{2ex} & Parking \\
         \multicolumn{2}{c}{\includegraphics[width=0.45\textwidth]{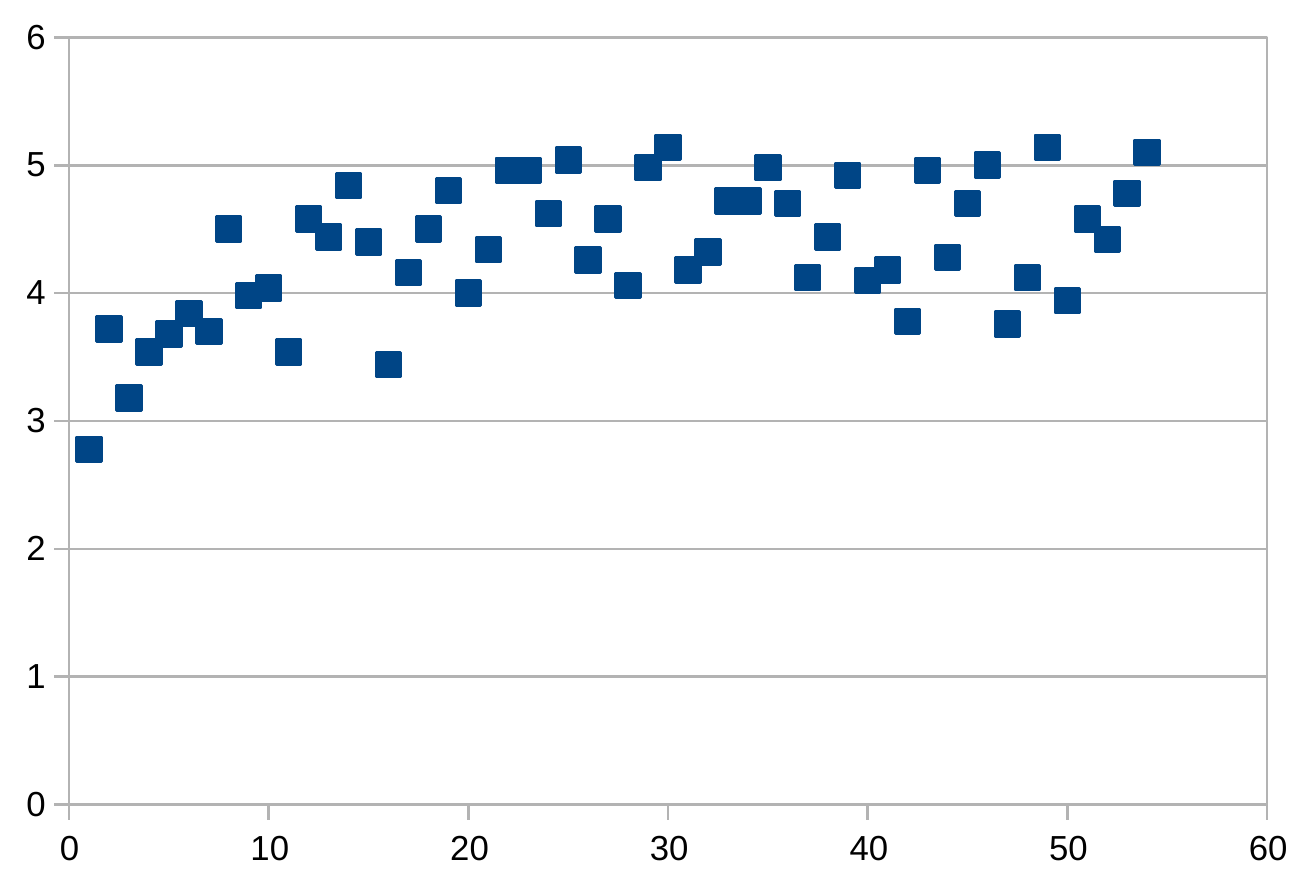}}\\
         \multicolumn{2}{c}{Transport\vspace{2ex}}\\
    \end{tabular}
    \caption{The average IPC score ($y$-axis) obtained on the training batch by the $n=50$ policies sampled at each iteration of the learning process ($x$-axis). Note that the variance is in large part due to random generation of the training batch (every iteration has a newly generated set of problems). The dynamic of the scores suggests that most of the policy improvement happens in the first handful of iterations.}
    \label{fig:training}
\end{figure*}

The training setup follows Algorithm \ref{alg:learning}, with the number of policies sampled at each iteration $n=50$ and the the number of elite policies $m=10$. The smoothing factor is $\alpha=0.7$, as suggested by \cite{Mannor2003}. For each of the domains, we train the model for 48 hours using 32 CPUs. Depending on the domain, this allows the training process to perform between 41 (\emph{Floortile}) and 91 (\emph{Elevators}) iterations. However, most of the policy improvement seems to happen in the first several iterations of the training process. This can be seen in Figure \ref{fig:training}. The plots show the average IPC score obtained by the policies sampled at each iteration. Note that the policies themselves are not the only source of variability: the training problems are randomly sampled for each iteration and some of them turn out to be harder than others despite using the same generator parameters. Even with this caveat, the plots suggest that most of the progress is made in the early iterations. In the extreme case of \emph{Floortile}, it is not clear from the plot whether the learner makes any progress at all. However, the results discussed in the next section suggest that a reasonably effective policy is still learned.

\subsection{Quantitative evaluation}

We perform two separate evaluations of the trained policies. For the first one we use the actual problem sets from the satisficing track of IPC 2011. For every domain and planner configuration we perform 10 test runs on the problem set and average the scores. In the second evaluation, we use unseen problems generated in the same manner as the training ones. The motivation for the second approach is twofold. First, it allows for performing each of the 10 test runs on a different set of problems and in doing so allows us to assess the planners over a total of 200 rather than just 20 problems. Second, by removing any manual filtering of the problems, it allows us to observe the planners performance on problems unseen in training, but still exactly following the training distribution.

We compare the learned policies against four fixed search routines making use of one of the search techniques available to the parametrized planner. For greedy best-first search we use the Fast Downward's implementation. The other baselines are implemented by setting the parameters of our parametrized planner accordingly. Overall the set of baseline planners includes:
\begin{itemize}
    \item plain greedy BFS [GBFS];
    \item $\epsilon$-greedy BFS with $\epsilon=0.5$ ($R=0$, $c=0$) [$\epsilon$-greedy];
    \item greedy BFS following node expansion with $R=5$ random walks of length $L=10$ on condition that the lowest known heuristic value $h_{min}$ has not changed for last $S=10$ expansions ($c=0$) [RW];
    \item local BFS with a span of $C=200$ expansions ($R=0$, $c=1$) [Local];
    \item a combination of all of the above, instantiating the parametrized planner template with $\epsilon=0.5$, $S=10$, $R=5$, $L=10$, $C=200$ and $c=0.5$ [Mixed].
\end{itemize}

The motivation behind the \emph{mixed} configuration is to assess performance of an algorithm making a balanced use of all the available techniques. In this configuration the planner interleaves between 100 expansions using the global open list and 100 expansions of local search. It follows the expansions with random walks if the search has not progressed for the last 10 expansions. A random node is selected for expansion instead of the one with lowest heuristic value with the probability of 0.5.

We remark that a direct comparison of our system against recent related work \cite{Gomoluch2019} would be difficult because of its requirement to train on substantially smaller problems under time limits of the order of 5 seconds.

The IPC scores obtained by each of the considered planners are reported in Table \ref{tab:ipc-sets} (IPC problem sets) and Table \ref{tab:test-sets} (generated test sets). The learned policies score higher than any of the baselines in all cases except \emph{No-mystery} set from IPC, where it narrowly trails plain greedy BFS. In general, in \emph{No-mystery} performance of greedy and $\epsilon$-greedy search and the learned policy is very similar. In \emph{Transport} and \emph{Parking} the baseline reaching performance closest to the learner is local search.
Importantly, the \emph{default} configuration obtains the lowest overall scores, clearly showing the advantage of a learned policy over a naive combination of all the search techniques.

\begin{table}
\begin{tabular}{|l|lllll|l|}
\hline
                  & E              & F            & N             & P              & T             & Sum            \\
\hline
GBFS              & 16.15          & 4.17         & \textbf{8.84} & 8.08           & 0             & 37.24          \\
$\epsilon$-greedy & 13.64          & 6.05         & 8.43          & 5.91           & 0             & 34.03          \\
RW                & 14.77          & 2.35         & 6.58          & 6.59           & 0.52          & 30.81          \\
Local             & 15.36          & 4.55         & 5.79          & 10.9           & 1             & 37.6           \\
Mixed             & 10.98          & 2.87         & 7.49          & 5.5            & 0.45          & 27.29          \\
Learned           & \textbf{16.71} & \textbf{6.5} & 8.56          & \textbf{11.77} & \textbf{1.11} & \textbf{44.64} \\
\hline
\end{tabular}
\label{tab:ipc-sets}
\caption{IPC scores for IPC problem sets (average of 10 runs). \emph{Elevators} (E), \emph{Floortile} (F), \emph{No-mystery} (N), \emph{Parking} (P) and \emph{Transport} (T).}
\end{table}

\begin{table}
\centering
\begin{tabular}{|l|lllll|l|}
\hline
                  & E              & F             & N             & P              & T            & Sum            \\
\hline
GBFS              & 14.69          & 2.27          & 8.19          & 9.3            & 2.6          & 37.05          \\
$\epsilon$-greedy & 13.09          & 2.68          & 8.95          & 7.49           & 2.75         & 34.97          \\
RW                & 14.68          & 0.47          & 6.78          & 8.07           & 3.75         & 33.75          \\
Local             & 15.99          & 1.93          & 7.16          & 12.23          & 4.67         & 41.97          \\
Mixed             & 11.65          & 1.26          & 6.7           & 6.22           & 3.01         & 28.84          \\
Learned           & \textbf{16.42} & \textbf{3.31} & \textbf{9.06} & \textbf{13.36} & \textbf{5.3} & \textbf{47.44} \\
\hline
\end{tabular}
\label{tab:test-sets}
\caption{IPC scores for randomly generated test problems (average over 10 sets). \emph{Elevators} (E), \emph{Floortile} (F), \emph{No-mystery} (N), \emph{Parking} (P) and \emph{Transport} (T).}
\end{table}

\subsection{Qualitative evaluation}

In this section, we examine the search policies learned for each of the domains. We do so by recording the search parameters designated by the policies while solving selected validation problems and interpreting the resulting search behaviour.

\subsubsection{Elevators} In this domain, the easiest problems are solved with GBFS ($c=0$, $\epsilon \approx 0.01$). A large number ($>15$) of relatively short random walks (3 steps) is used when the search fails to progress for around $30$ node expansions. In larger instances, the value of $\epsilon$ is even smaller ($\epsilon \approx 0.001$), the random walks get longer (9 steps) and there is more of them ($>30$). This makes intuitive sense in the logistics-style domain of \emph{Elevators}: random walks may increase the chance of finding a solution but also increase its cost. Consequently, long random walks are better avoided when the problem is likely to be solved in time without them. On the other hand, they are useful when the time limit becomes an issue. For large problems, a small number of local expansions is performed towards the end of the search (2 in a cycle of 50), but it is not clear whether this has substantial effect on the results.

\subsubsection{Floortile} In \emph{Floortile} the policy results mostly in $\epsilon$-greedy BFS. It starts with a very high value of $\epsilon \approx 0.8$, which means that the heuristic value is ignored most of the time. This is not entirely surprising given the performance of baseline planners guided by the same heuristic. For the largest problems solved within the time limit, the value of $\epsilon$ gradually falls towards the end of the search. The policy performs better than the baselines, but still fails to solve most of the problems.

\subsubsection{No-mystery} The values of $\epsilon$ in \emph{No-mystery} are even more extreme than in \emph{Floortile}, ranging between 0.75 and 0.98 and increasing as $h_{min}$ decreases. Neither random walks nor local search are used. On the one hand, the policy shows how a \emph{delete relaxation} heuristic is of limited use. On the other hand, it also brings up a limitation of the parametrized search algorithm in its current form. While the value of the heuristic is virtually unused, it still needs to be computed for every encountered state, for the purpose of ordering the open list. We aim to address this issue in future work. 

\subsubsection{Parking} In small \emph{Parking} instances the policy's behaviour is close to best-first search. Interestingly, this is achieved by setting the cycle length $C$ to a very small value (a cycle with 0 or 1 local expansion is equivalent to global search). For larger instances, this approach is mixed with evenly balanced cycles of several global and local expansions. The $\epsilon$ values are typically around 0.1. 

\subsubsection{Transport} On smaller \emph{Transport} problems the learned policy's behaviour is close to global $\epsilon$-greedy BFS, with the value $\epsilon$ ranging between 0.08 and 0.2. This is achieved by using a very short cycle with no local expansions. In larger instances the policy performs iterated local search of varying span. The number of expansions in a single cycle is typically varies through the search, between the values of 10 and 150. Extensive use of local search makes intuitive sense in the logistics-style \emph{Transport} domain, where focusing on a subset of the search space can often allow for faster progress. Occasionally, the search approaches global BFS by decreasing $C$ to very low values. The value of $\epsilon$ increases as the search progresses, up to about 0.25.

\section{Conclusion}

We have introduced a parametrized search algorithm flexibly combining multiple search techniques in a way determined by the values of its parameters. We further constructed a simple neural policy model for designating the values of the algorithm's parameters given a high-level representation of the current state of the search. Using the cross-entropy method we have trained the model on problems from five different planning domains. The empirical evaluation shows that the learner is able to discover effective distribution-specific search policies.

Besides the strengths of the approach, the empirical evaluation has also shown some of its limitations. While the current approach has the capacity to ignore the value of a heuristic it finds misleading, it can neither replace it nor stop computing it for the all encountered states, even if the value is not going to be used. One possible solution would be to make to choice of heuristic subject to the learning process, for example by allowing the algorithm to work with multiple open lists. It is also be possible to extend the algorithm with other state-of-the-art techniques from classical planning, such as \emph{preferred operators} or novelty-based search \cite{Lipovetzky2017}.

Another direction of future work is the representation of the planner's state. Capturing the relevant information in a domain-independent manner is a huge challenge and the key to improve the expressiveness of the learned policies.

\bibliographystyle{aaai}
\bibliography{references}

\end{document}